\documentclass{article}

\usepackage{arxiv}

\usepackage[utf8]{inputenc} % allow utf-8 input
\usepackage[T1]{fontenc}    % use 8-bit T1 fonts
\usepackage{hyperref}       % hyperlinks
\usepackage{url}            % simple URL typesetting
\usepackage{booktabs}       % professional-quality tables
\usepackage{amsfonts}       % blackboard math symbols
\usepackage{nicefrac}       % compact symbols for 1/2, etc.
\usepackage{microtype}      % microtypography
\usepackage{lipsum}
\usepackage{graphicx}
\graphicspath{ {./images/} }
\usepackage{bm}
 \usepackage{amsmath}
 \usepackage{overpic}
 \usepackage{array}
 \usepackage{multirow}
 \usepackage{booktabs}
\usepackage{tabularx}
\usepackage{tabularray}
\usepackage[table]{xcolor}
\usepackage{color,colortbl}
\definecolor{LightBlue}{rgb}{0.88,0.9,0.9}
\colorlet{LightBlue}{LightBlue!40}
\usepackage{natbib}
\usepackage{subcaption}

\title{MAGIC-Flow: Multiscale Adaptive Conditional Flows for Generation and Interpretable Classification}

\author{
 Luca Caldera \\
  MOX, Department of Mathematics \\
  Politecnico di Milano \\
  Milan, Italy \\
  \texttt{luca.caldera@polimi.it} \\
   \And 
 Giacomo Bottacini \\
  MOX, Department of Mathematics \\
  Politecnico di Milano \\
  Milan, Italy \\
  \texttt{giacomo.bottacini@polimi.it} \\
   \And
 Lara Cavinato \\
  MOX, Department of Mathematics \\
  Politecnico di Milano \\
  Milan, Italy \\
  \texttt{lara.cavinato@polimi.it} \\
  \And 
 For the Alzheimer’s Disease Neuroimaging Initiative\thanks{%
Data used in preparation of this article were obtained from the Alzheimer’s
Disease Neuroimaging Initiative (ADNI) database (\url{adni.loni.usc.edu}). 
As such, the investigators within ADNI contributed to the design and implementation 
of ADNI and/or provided data but did not participate in analysis or writing of this report. 
A complete listing of ADNI investigators can be found at: 
\url{http://adni.loni.usc.edu/wp-content/uploads/how_to_apply/ADNI_Acknowledgement_List.pdf}.
} \\
}

\begin{document}
\maketitle
\begin{abstract}
Generative modeling has emerged as a powerful paradigm for representation learning, but its direct applicability to challenging fields like medical imaging remains limited: mere generation, without task alignment, fails to provide a robust foundation for clinical use. We propose MAGIC-Flow, a conditional multiscale normalizing flow architecture that performs generation and classification within a single modular framework. The model is built as a hierarchy of invertible and differentiable bijections, where the Jacobian determinant factorizes across sub-transformations. We show how this ensures exact likelihood computation and stable optimization, while invertibility enables explicit visualization of sample likelihoods, providing an interpretable lens into the model’s reasoning. By conditioning on class labels, MAGIC-Flow supports controllable sample synthesis and principled class-probability estimation, effectively aiding both generative and discriminative objectives. We evaluate MAGIC-Flow against top baselines using metrics for similarity, fidelity, and diversity. Across multiple datasets, it addresses generation and classification under scanner noise, and modality-specific synthesis and identification. Results show MAGIC-Flow creates realistic, diverse samples and improves classification.
MAGIC-Flow is an effective strategy for generation and classification in data-limited domains, with direct benefits for privacy-preserving augmentation, robust generalization, and trustworthy medical AI.
\end{abstract}

% keywords can be removed

%\begingroup
%\renewcommand\thefootnote{}%
%\footnotetext{
%Data used in preparation of this article were obtained from the Alzheimer’s
%Disease Neuroimaging Initiative (ADNI) database (\url{adni.loni.usc.edu}). 
%As such, the investigators within ADNI contributed to the design and implementation 
%of ADNI and/or provided data but did not participate in analysis or writing of this report. 
%A complete listing of ADNI investigators can be found at: 
%\url{http://adni.loni.usc.edu/wp-content/uploads/how_to_apply/ADNI_Acknowledgement_List.pdf}.
%}%
%\addtocounter{footnote}{0}%
%\endgroup

\section{Introduction}
\label{sec:intro}

Generative modeling has become a cornerstone of modern machine learning, powering advances in representation learning, data augmentation, and controllable synthesis. In addition to semi-supervised and transfer learning, generative models have been increasingly used to mitigate data scarcity. GANs \citep{goodfellow2014generative, makhlouf2023use, han2018gan}, VAEs \citep{sohn2015learning, diamantis2022endovae}, and diffusion models \citep{nichol2021improved, dhariwal2021diffusion, hung2023med, kazerouni2022diffusion} have been proven effective at synthesizing realistic samples and improving classifier robustness \citep{shorten2019survey, shin2018medical}.  
Yet, in sensitive domains such as medical imaging, their limitations are clear. Datasets are small, expensive to curate, and biased by acquisition protocols \citep{zech2018variable}. Models that generate realistic images without offering \emph{task alignment} or \emph{interpretability} fall short in clinical contexts, where reliability and transparency are critical. Moreover, adversarial and diffusion-based models suffer from mode collapse, hallucination, and instability \citep{arjovsky2017wasserstein, cohen2018distribution, yi2019generative}, undermining their trustworthiness.  

Normalizing flows provide a principled alternative. Unlike GANs or diffusion models, flows offer \emph{exact likelihood estimation, stable training, and invertible mappings}, making them attractive for data-limited, safety-critical applications. However, existing architectures such as RealNVP \citep{dinh2016density} and Glow \citep{kingma2018glow} have been designed almost exclusively for \emph{generation}. When adapted to classification, they typically rely on auxiliary discriminative heads or downstream training on latent embeddings. This separation prevents a unified treatment of generation and classification and limits interpretability, as likelihoods are not directly exploited for decision-making.  

In this context, we introduce \textbf{MAGIC-Flow}, a conditional multiscale normalizing flow that unifies \emph{generation and classification} within a shared invertible architecture, standing out from canonical flows, which remain purely generative, and from hybrid approaches that bolt on classification heads. Our key contribution is to show that with only \emph{minor architectural changes}, the same flow backbone can be adapted to both tasks:  
\begin{itemize}
    \item In its generative configuration, MAGIC-Flow synthesizes diverse, controllable samples with exact likelihoods.  
    \item In its discriminative configuration, MAGIC-Flow leverages the same invertible mappings to derive class probabilities directly from normalized densities, avoiding opaque embeddings or external classifiers.  
\end{itemize}  

% We benchmark MAGIC-Flow against state-of-the-art generative models (SNGAN \citep{miyato2018spectral}, StyleGAN2-DiffAug-LeCam \citep{karras2020analyzing, zhao2020differentiable, tseng2021regularizing}, ADC-GAN \citep{hou2022conditional, odena2017conditional}, DDPM \citep{nichol2021improved, dhariwal2021diffusion}, CVAE \citep{sohn2015learning}) and discriminative baselines (CNNs pretrained on RadImageNet \citep{mei2022radimagenet}, ViTs \citep{dosovitskiy2020image, jain2024comparative, liu2021swin, he2023transformers}). Unlike these approaches, MAGIC-Flow does not treat the two tasks as independent; instead, it shows that a \emph{single invertible backbone} suffices for both, while preserving tractable likelihoods and stability.  

We benchmark MAGIC-Flow against state-of-the-art generative models: unlike these approaches, MAGIC-Flow does not treat the two tasks as independent; instead, it shows that a \emph{single invertible backbone} suffices for both, while preserving tractable likelihoods and stability.  
The evaluation is conducted on two clinically relevant challenges: (i) generation and classification under scanner noise, where classes are defined by acquisition artifacts rather than semantics - which represents a more challenging task with respect to e.g. diagnosis prediction -, and (ii) cross-modality generalization across imaging modalities. In both settings, MAGIC-Flow consistently improves over baselines in terms of \emph{sample fidelity, diversity, and classification accuracy}, while uniquely offering \emph{inherent interpretability via likelihood visualization}.  

\paragraph{Contributions.} Our main contributions are threefold:
\begin{itemize}
    \item We propose MAGIC-Flow, the first conditional multiscale normalizing flow that supports both generation and classification within a shared invertible framework.  
    \item We show that only minor architectural changes are required to switch between the two tasks, while maintaining exact likelihood computation and interpretability.  
    \item We demonstrate that MAGIC-Flow outperforms strong generative and discriminative baselines in challenging medical imaging settings, providing a principled and trustworthy foundation for aiding clinically relevant AI.  
\end{itemize}

\section{Theoretical Foundation}
\label{sec:theory}
Normalizing flows are a class of generative models that represent complex probability distributions by transforming a simple base distribution through a sequence of invertible and differentiable mappings. Let $z \sim p_Z(z)$ be a latent variable and $f$ an invertible mapping. A flow defines $x = f(z)$ and $z = f^{-1}(x)$. The inverse mapping $x \mapsto z$ enables tractable likelihood evaluation, while the forward mapping $z \mapsto x$ enables exact sampling. By the change-of-variables formula, the log-likelihood of $x$ is
\begin{equation}
\log p_X(x) = \log p_Z(f^{-1}(x)) + \log \left| \det \frac{\partial f^{-1}(x)}{\partial x} \right|.
\end{equation}

The design of flow transformations balances two requirements: (i) \emph{expressiveness}, to capture complex distributions, and (ii) \emph{tractability}, to allow efficient computation of inverses and Jacobian determinants. Several architectures embody these trade-offs: NICE \citep{dinh2014nice} introduced additive coupling layers; RealNVP \citep{dinh2016density} extended this with affine couplings and multiscale architectures; Glow \citep{kingma2018glow} further improved expressiveness with invertible $1 \times 1$ convolutions. See \cite{kobyzev2020normalizing, papamakarios2021normalizing} for surveys.

Normalizing flows therefore provide a principled framework for generative modeling, combining exact likelihood training with efficient sampling. We now extend this formulation to the \emph{conditional} setting, which is central to our architecture.

\paragraph{Conditional normalizing flows.} 
Conditional normalizing flows (cNFs) incorporate auxiliary information $y$ (e.g., class labels or embeddings) into the transformation $f(\cdot, y)$, enabling modeling of conditional densities $p_{X \mid Y}$. Applications include structured prediction \citep{winkler2019learning}, guided image generation with conditional invertible neural networks (cINNs) \citep{ardizzone2019guided}, and autoregressive conditional flows (CAFLOW) \citep{batzolis2021caflow}. More recent works apply cNFs to mitigate mode collapse \citep{kanaujia2024advnf} or enable anomaly detection \citep{gudovskiy2022cflow}. 

Our theoretical contribution is to show that the two cornerstone properties of flows—\emph{invertibility} and \emph{Jacobian factorization}—also hold in the conditional setting. Extensive proofs are provided in Appendix \ref{app:conditional-cov} and \ref{app:inv_transf}. 

\paragraph{Invertibility property.} Under mild assumptions, the conditional change-of-variable formula is
\begin{equation}
p_{X \mid Y}(x \mid y) = p_Z\!\left(f^{-1}(x, y)\right) \cdot 
\left| \det \frac{\partial f^{-1}(x, y)}{\partial x} \right|.
\label{eq:cond_cov}
\end{equation}
Here, $x \in \mathbb{R}^{C \times H \times W}$ denotes images, $y \in \mathbb{R}^K$ encodes labels (e.g., one-hot or learned embeddings), and $z \sim \mathcal{N}(0,I)$ is the latent representation. Conditioning is handled by parameterizing each flow transformation $f_i(\cdot,y)$ with $y$, while preserving bijectivity in $x$.

\paragraph{Factorization property.}
As in the unconditional case, the overall flow can be expressed as a composition of $N$ bijective mappings $f = f_N \circ f_{N-1} \circ \ldots \circ f_1$ with intermediate states $h_i = f_i(h_{i-1}, y)$, where $h_0 = x$ and $h_N = z$. Since each $f_i(\cdot,y)$ is bijective in $x$, their composition is bijective as well. The Jacobian determinant factorizes across layers:
\begin{equation}
    \left| \det \frac{\partial f^{-1}(x, y)}{\partial x} \right|
    = \prod_{i=1}^N \left| \det \frac{\partial f_i^{-1}(h_i, y)}{\partial h_i} \right|.
\end{equation}

This result establishes that conditional flows inherit the same tractability guarantees as unconditional flows, while enabling controlled generation and likelihood-based classification. It forms the theoretical foundation of MAGIC-Flow, which builds upon conditional invertibility and factorization to support both generative and discriminative tasks within a shared architecture.

\section{Unified Architecture of MAGIC-Flow}
\label{sec:architecture}

Building on the conditional formulation from the previous section, we present the unified architecture of MAGIC-Flow. The theoretical foundation showed that invertibility and Jacobian factorization extend naturally to the conditional case, guaranteeing that conditional flows remain tractable. MAGIC-Flow instantiates these principles in a hierarchical, multiscale design that combines conditional flow steps, squeeze operations, and split operations. This framework is fully general: only the definition of the affine coupling transformation is task-specific, as discussed later in Section~\ref{sec:task_coupling}.

%\paragraph{Flow steps.} 
%A single conditional flow step (Figure~\ref{fig:model_architecture}a) consists of three standard invertible components:
%\begin{enumerate}
%    \item \textbf{ActNorm} \citep{kingma2018glow}: a channel-wise affine transformation initialized from data statistics \citep{salimans2016weight} and subsequently learned as trainable parameters.
%    \item \textbf{Invertible $1\times1$ Convolution} \citep{kingma2018glow}: a learnable channel permutation that improves flexibility while preserving invertibility and efficient Jacobian computation.
%    \item \textbf{Affine Coupling Transformation} \citep{dinh2016density}: the core operation, here extended to incorporate conditioning on $y$. The scale and shift functions of the coupling transformation are defined by task-specific layers (Section~\ref{sec:task_coupling}).
%\end{enumerate}

\begin{figure}[ht]
\centering
\begin{overpic}[width=0.85\linewidth]{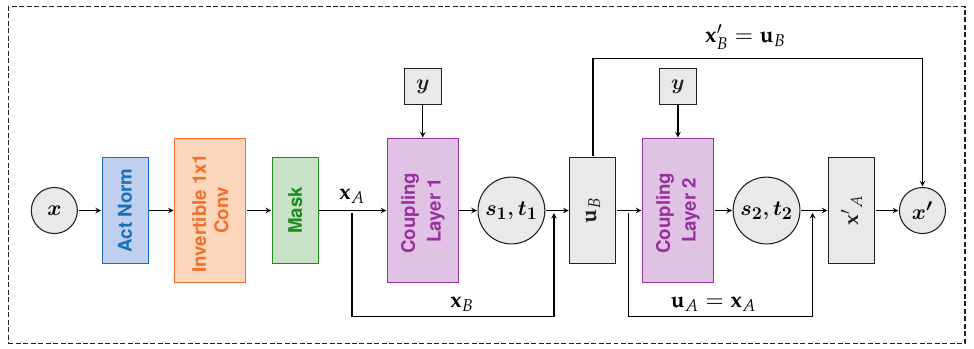}
    \put(108,18){\normalsize (a)}
\end{overpic}

\vspace{0.9em} 

\begin{overpic}[width=0.95\linewidth]{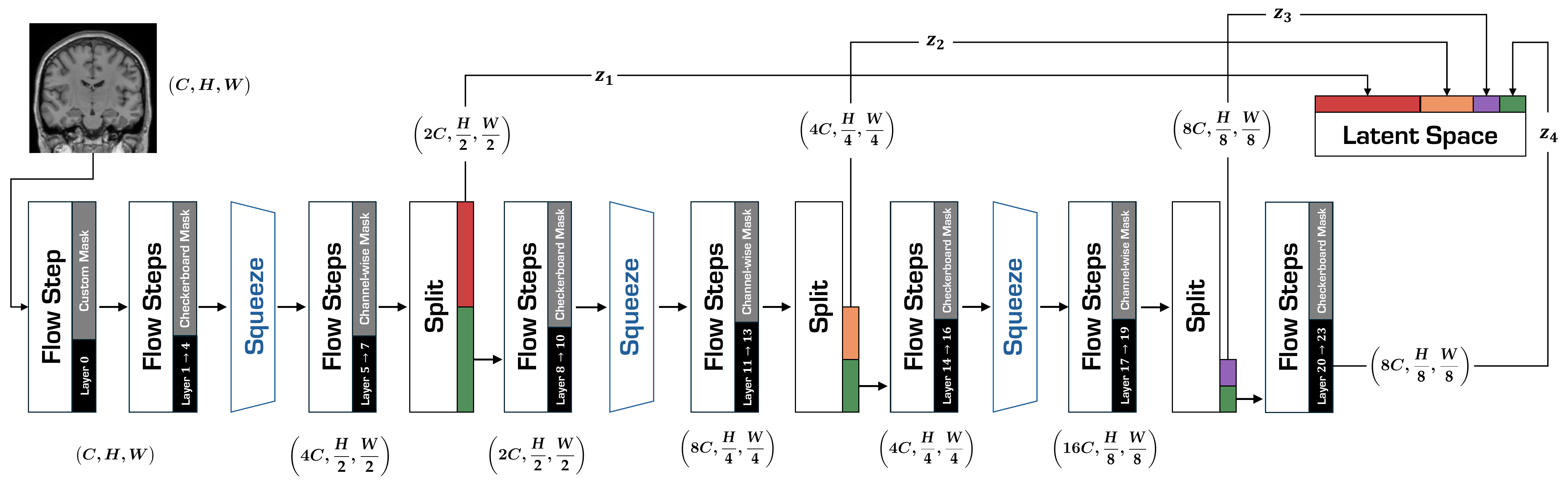}
    \put(101, 17){\normalsize (b)}
\end{overpic}

\caption{Illustration of components of the model: (a) single flow step and (b) full multiscale flow.}
\label{fig:model_architecture}
\end{figure}

\paragraph{Flow steps.} 
A single conditional flow step (Figure~\ref{fig:model_architecture}a) consists of three standard invertible components: (1) \textbf{ActNorm} \citep{kingma2018glow}: a channel-wise affine transformation initialized from data statistics \citep{salimans2016weight} and subsequently learned as trainable parameters; (2) \textbf{Invertible $1\times1$ Convolution} \citep{kingma2018glow}: a learnable channel permutation that improves flexibility while preserving invertibility and efficient Jacobian computation; (3) \textbf{Affine Coupling Transformation} \citep{dinh2016density}: the core operation, here extended to incorporate conditioning on $y$. The scale and shift functions of the coupling transformation are defined by task-specific layers (Section~\ref{sec:task_coupling}).

Concretely, the input \(\mathbf{x} \in \mathbb{R}^{C \times H \times W}\) is partitioned into \(\mathbf{x}_A\) and \(\mathbf{x}_B\) with a binary mask \(\mathbf{M}\), such that \(\mathbf{x}_A = \mathbf{M} \odot \mathbf{x}\) and \(\mathbf{x}_B = (1-\mathbf{M}) \odot \mathbf{x}\). The two partitions are alternately transformed, conditioned on the other partition and the label $y$, via scale and shift functions $\mathbf{s}_i, \mathbf{t}_i$:
\[
\mathbf{u}_A = \mathbf{x}_A, \quad
\mathbf{u}_B = \mathbf{x}_B \odot \exp(\mathbf{s}_1(\mathbf{x}_A, y)) + \mathbf{t}_1(\mathbf{x}_A, y),
\]
\[
\mathbf{x}'_A = \mathbf{u}_A \odot \exp(\mathbf{s}_2(\mathbf{u}_B, y)) + \mathbf{t}_2(\mathbf{u}_B, y), \quad
\mathbf{x}'_B = \mathbf{u}_B.
\]
The output is recombined as $\mathbf{x}' = \mathbf{x}'_A \oplus \mathbf{x}'_B$, with log-determinant
\[
\log|\det J| = \sum_{c,h,w}(1-M_{c,h,w})\,s_{1,c,h,w} 
+ \sum_{c,h,w}M_{c,h,w}\,s_{2,c,h,w}.
\]

\paragraph{Squeeze and split operations.} 
To model multiscale structure efficiently, MAGIC-Flow incorporates two standard architectural operations: \textit{Squeeze} that reshapes a tensor of size $(C,H,W)$ into $(4C, H/2, W/2)$, reducing spatial resolution while increasing channel depth. This allows coupling layers to operate over larger receptive fields without added convolutional cost \citep{hoogeboom2019emerging}; \textit{Split} that divides the feature map along the channel dimension into two halves. One half is factored out directly into the latent representation, while the other continues through subsequent flow steps. Splits yield a hierarchical latent variable $z$ that encodes fine-grained details at early stages and coarser structure at later stages \citep{ardizzone2019guided, gudovskiy2022cflow}.

\paragraph{Masking strategy.} 
To ensure that all dimensions of $\mathbf{x}$ are transformed across the flow, MAGIC-Flow alternates between three types of binary masks $\mathbf{M} \in \{0,1\}^{C \times H \times W}$: (1) \textbf{Checkerboard masks:} update alternating spatial locations to promote pixel-level mixing; (2) \textbf{Channel-wise masks:} transform subsets of channels, enabling feature-level transformations; (3) \textbf{Application-specific masks:} selectively emphasize semantically relevant regions, adapting the transformation to the downstream objective.
Alternating checkerboard and channel-wise masks across flow steps balances spatial and feature-level expressiveness while maintaining tractable Jacobian computations. See Appendix \ref{app:mask_design} for details and visualization.

\paragraph{Overall design.} 
%The full MAGIC-Flow architecture (Figure~\ref{fig:model_architecture}b) stacks 24 conditional flow steps organized in a multiscale hierarchy. The sequence begins with three checkerboard flow steps, followed by three channel-wise flow steps, after which a squeeze operation is applied. Subsequent blocks repeat this pattern: three checkerboard steps, followed by a split, then three channel-wise steps, another squeeze, and so on. At each split, half of the channels are factored into the latent representation, while the remainder is further transformed. This recursive decomposition produces a hierarchical latent representation that captures both local detail and global structure, while preserving tractable likelihood evaluation.
The full MAGIC-Flow architecture (Figure~\ref{fig:model_architecture}b) stacks 24 conditional flow steps organized in a multiscale hierarchy. The sequence begins with a flow step using an application-specific mask, followed by three checkerboard flow steps, a squeeze operation, and three channel-wise flow steps, after which a split operation is applied. Subsequent blocks repeat this pattern: three checkerboard steps, a squeeze, three channel-wise steps, and another split, and so on. At each split, half of the channels are factored into the latent representation, while the remainder is further transformed. This recursive decomposition produces a hierarchical latent representation that captures both local detail and global structure, while preserving tractable likelihood evaluation.

MAGIC-Flow is a unified framework where the conditional invertibility and factorization properties are instantiated through a hierarchy of flow steps, squeeze and split operations, and structured masking. This design ensures both expressive generative modeling and efficient discriminative feature learning. 
In this way, MAGIC-Flow serves as both a conditional generative model and a conditional density estimator for classification, within a single framework.
The only task-specific component lies in the definition of the affine coupling layers, which we detail next in Section~\ref{sec:task_coupling}.

\section{Task-Specific Coupling Layers and Learning Objectives}
\label{sec:task_coupling}

While the unified architecture of MAGIC-Flow (Section~\ref{sec:architecture}) guarantees conditional invertibility and tractable likelihood computation, its flexibility stems from the design of the \emph{task-specific affine coupling layers}. These layers specialize the shared framework for two complementary objectives: high-fidelity conditional image generation and accurate label-informed classification.

\subsection{Task-Specific Affine Coupling Layers}

Both coupling layers adopt a convolutional conditioning network that maps the input $\mathbf{x} \in \mathbb{R}^{C \times H \times W}$ and label $\mathbf{y} \in \mathbb{R}^K$ into scale and shift parameters $(s,t)$. The key difference lies in how conditioning is injected and how feature transformations are optimized for each task. See Appendix~\ref{app:aff_cou_arch} for detailed architectures of both layers.

\paragraph{Generation coupling layer.} 
For conditional synthesis, the coupling network prioritizes \emph{expressiveness} and \emph{contextual richness}. Conditioning information is injected at multiple depths via FiLM layers \citep{perez2018film}, while CBAM modules \citep{woo2018cbam} provide channel- and spatial-wise attention. Residual blocks with FiLM modulation \citep{he2016deep} enhance representational capacity. To capture long-range dependencies and multi-scale context, we integrate global context blocks \citep{hu2018squeeze, cao2019gcnet} and ASPP-SE modules \citep{chen2017deeplab, hu2018squeeze}. Together, these components enable the coupling transformation to leverage label information across scales, producing condition-aware transformations richer and more flexible than conventional shallow coupling networks \citep{dinh2016density, kingma2018glow}.

\paragraph{Classification coupling layer.} 
For discriminative modeling, the coupling network is streamlined to emphasize \emph{label-informed, discriminative transformations} rather than multi-scale expressiveness. It includes a residual label embedding module, FiLM-based conditioning, CBAM attention, and residual convolutional blocks. Convolutional features are normalized, regularized with dropout, and then fused with the label embedding. Finally, parallel prediction heads output translation and bounded scale parameters. This design ensures that transformations are explicitly conditioned on label information, yielding features that are discriminative and robust for likelihood-based classification.

\subsection{Generation and Classification Procedures}
\label{subsec:generation_classification}

With the task-specific couplings defined, we now describe how MAGIC-Flow performs the two primary tasks: image generation and classification. Both tasks leverage the same invertible backbone, differing only in the coupling transformation.

\paragraph{Training objective.}
Given a dataset of paired inputs and labels $\{(x_i, y_i)\}_{i=1}^N$, the parameters of MAGIC-Flow are optimized by maximizing the conditional log-likelihood:
\[
\mathcal{L}(\theta) = \frac{1}{N} \sum_{i=1}^N \log p_{X \mid Y}(x_i \mid y_i)
= \frac{1}{N} \sum_{i=1}^N \Big[ \log p_Z(f_\theta^{-1}(x_i, y_i)) + \log \left| \det \frac{\partial f_\theta^{-1}(x_i, y_i)}{\partial x_i} \right| \Big].
\]

\paragraph{Generation.}  
To synthesize an image conditioned on a label $\mathbf{y}$, we draw a latent variable $\mathbf{z} \sim \mathcal{N}(0,I)$ and apply the inverse flow: $\mathbf{x} = f^{-1}(\mathbf{z}, \mathbf{y})$. The hierarchical architecture (squeeze, split, and multiscale flow steps) ensures that both global structure and fine detail are generated consistently. The generation coupling layers inject label information at every stage, enabling the synthesis of diverse and realistic images aligned with $\mathbf{y}$.

\paragraph{Classification.}  
To classify an input $\mathbf{x}$, MAGIC-Flow evaluates conditional likelihoods. The forward flow maps the input into the latent space $\mathbf{z} = f(\mathbf{x}, \mathbf{y})$ and the conditional log-likelihood can be computed using Equation \ref{eq:cond_cov}. At inference time, the predicted label $\hat{\mathbf{y}}$ is obtained by maximum likelihood:
\[
\hat{\mathbf{y}} = \arg\max_{\mathbf{y}} \log p_{X \mid Y}(\mathbf{x} \mid \mathbf{y}).
\]

% Importantly, both tasks are performed within the same conditional flow architecture. The shared backbone guarantees tractable likelihoods and invertible mappings, while the task-specific coupling layers adapt the transformation to either maximize generative fidelity or enhance discriminative power. This unification allows MAGIC-Flow to serve as both a conditional generative model and a conditional density estimator for classification, within a single framework.

\section{Interpretability via Likelihood Attribution Maps} 
\label{sec:heatmaps}

A key advantage of MAGIC-Flow over existing generative frameworks is its ability to compute \emph{exact likelihoods}. This property enables not only training via maximum likelihood estimation, but also principled interpretability through likelihood decomposition. We introduce \emph{likelihood attribution maps}, which quantify the spatial contribution of each image location to the conditional log-likelihood. Unlike heuristic, post-hoc methods such as Grad-CAM \citep{selvaraju2017grad} or integrated gradients \citep{sundararajan2017axiomatic}, our attribution maps are derived directly from the internal computations of the flow, ensuring faithfulness and model consistency.
To compute $\mathcal{H}(\mathbf{x}, \mathbf{y}) \in \mathbb{R}^{C \times H \times W}$, we traverse the model as follows:

\begin{enumerate}
    \item \textbf{Forward pass:} Map a input $\mathbf{x}$ conditioned on $\mathbf{y}$ to the latent representation $\mathbf{z}$ via the conditional flow.

    \item \textbf{Backward pass:} Accumulate contributions to the log-likelihood from two sources:
    \begin{itemize}
        \item \emph{Local Jacobian contributions:} Each affine coupling layer $i$ contributes
        \[
        \ell_J^i(\mathbf{x},\mathbf{y})_{c,h,w} = (1-M_{c,h,w})\, s_{1,c,h,w}(\mathbf{x}_A,\mathbf{y}) + M_{c,h,w}\, s_{2,c,h,w}(\mathbf{u}_B,\mathbf{y}).
        \]
        \item \emph{Latent contributions:} In the multiscale architecture, split operations factor out subsets $\mathbf{z}_j$ ($j=1,\dots,S+1$, where $S$ is the number of splits), each contributing via its Gaussian prior $\log p(\mathbf{z}_j)_{c,h,w}.$    
    \end{itemize}

    \item \textbf{Accumulation:} Starting from the final latent subset ($\mathbf{z}_4$ in Figure \ref{fig:model_architecture}b, size $(8C, H/8, W/8)$), we add its log-probability $\log p(\mathbf{z}_4)$ to the Jacobian contributions of the last flow layers ($i=20,\dots,23$), which share the same resolution. At the subsequent split, we concatenate this attribution map with the factored-out latent part ($\mathbf{z}_3$, also $(8C, H/8, W/8)$) along the channel dimension, yielding a combined attribution map of size $(16C, H/8, W/8)$. Continuing backward, we add the Jacobian contributions from layers $i=17,\dots,19$, then \emph{unsqueeze} the map to $(4C, H/4, W/4)$ to align with the earlier stage (see Figure \ref{fig:deriv_heatmaps} in Appendix \ref{app:likelihood} for a detailed illustration of these initial steps). This process of summation, concatenation, and reshaping is repeated until the full attribution map $\mathcal{H}(\mathbf{x}, \mathbf{y})$ is obtained.
\end{enumerate}

This recursive construction yields a faithful, spatially resolved map of how each pixel contributes to the conditional log-likelihood, providing insight into MAGIC-Flow’s reasoning while fully respecting the model’s invertible structure.

%\begin{figure}[t]
%    \centering
%    \includegraphics[width=0.9\linewidth]{iclr2026/model_imgs/explainability.png}
%    \caption{\textbf{Likelihood attribution maps.} Starting from latent factors (green) at the deepest level, contributions are traced backward through inverse split and squeeze operations, with Jacobian terms accumulated at each step. The process reconstructs a pixel-level attribution map $\mathcal{H}(\mathbf{x}, \mathbf{y})$, offering principled, likelihood-based interpretability of MAGIC-Flow’s predictions.}
%    \label{fig:explainability}
%\end{figure}

\section{Experiments}
\label{sec:experiments}

We evaluate MAGIC-Flow on two tasks: \emph{conditional generation} (scanner- and modality-conditioned) and \emph{classification} (scanner identification). Each task is tested on diverse, publicly available datasets, compared against strong baselines, and evaluated with task-appropriate metrics. All datasets and preprocessing details are provided in Section \ref{app:preproc} and \ref{app:datasets}.

\subsection{Generation Experiments}

\paragraph{Tasks and Datasets.}  
We assess two conditional generation tasks: (i) scanner-conditioned generation, evaluating whether MAGIC-Flow can synthesize realistic MRI slices that capture scanner-specific characteristics; and (ii) modality-conditioned generation, testing the ability to generate anatomically consistent images across MRI (T1, T2, FLAIR) and PET (FDG, amyloid, tau) modalities. Scanner-conditioned experiments use T1-weighted MRI slices from PPMI \citep{marek2011parkinson}, IXI \citep{ixidata}, and SALD \citep{wei2017structural}. Modality-conditioned experiments use multimodal MRI and PET from ADNI3/4 \citep{jack2008alzheimer}. Dataset statistics are summarized in Table~\ref{tab:datasets}a and~\ref{tab:datasets}b of Appendix \ref{app:datasets}.

\paragraph{Evaluation Metrics and Benchmarks.}  
We report FID and KID \citep{heusel2017gans, binkowski2018demystifying} with domain-adapted feature extractors (FID$_\text{Rad}$ and KID$_\text{Rad}$ pre-trained on RadImageNet; FID$_\text{SwAV}$ on SwAV) for medical realism \citep{mei2022radimagenet, caron2020unsupervised}. To capture sample-level fidelity and diversity, we use PRDC metrics (Precision, Recall, Density, Coverage) \citep{kynkaanniemi2019improved, naeem2020reliable}, and intra-/inter-class MS-SSIM \citep{wang2003multiscale, odena2017conditional}. Details are provided in Appendix \ref{subsubsec:distr_sim} and \ref{subsubsec:sampl_sim}. We compare against GANs (SNGAN \citep{miyato2018spectral}, StyleGAN2-DiffAug-LeCam \citep{karras2020analyzing, zhao2020differentiable, tseng2021regularizing}, ADC-GAN \citep{hou2022conditional}), diffusion models (DDPM \citep{dhariwal2021diffusion}), and latent-variable models (CVAE \citep{sohn2015learning}). See Appendix~\ref{appendix:benchmarks_gen} for details.

\subsection{Classification Experiments}

\paragraph{Task and Dataset.}  
We evaluate MAGIC-Flow on scanner classification, where the model discriminates between seven scanners, including scanners with comparable noise profiles e.g., Siemens Prisma vs. Prisma Fit, using unbalanced datasets to test robustness.
%\paragraph{Datasets.}  
Coronal slices from PPMI, SALD, IXI, and ADNI3 are used. Splits are performed across 5 folds with performance averaged for robustness. See Table~\ref{tab:datasets}a of Appendix \ref{app:datasets}.
% @Luca METTERE NEI SUPPLEMENTARY LE TABELLE DEI DATASET E IL PREPROCESSING (DA RIPRENDERE ANCHE NEL REPRODUCIBILITY STATEMENT.

\paragraph{Evaluation Metrics and Benchmarks.}  
We report Accuracy, Balanced Accuracy, AUC, and macro-averaged Precision, Recall, and F1-score, accounting for both overall and class-balanced performance. Details are provided in 
Appendix \ref{app:class_metrics}.
%\textbf{SUPPLEMENTARY MATERIALS/APPENDIX}.
%\paragraph{Benchmarks.}  
We compare against CNNs pretrained on RadImageNet \citep{mei2022radimagenet} (ResNet-50, DenseNet-121, InceptionV3, InceptionResNetV2) and Vision Transformers (ViT, ViT-ResNet, Swin Transformer) \citep{dosovitskiy2020image, jain2024comparative, liu2021swin}. Baselines are evaluated with and without pretraining. See Appendix~\ref{appendix:benchmarks_class} for details.

% \subsection{Datasets and Preprocessing}
% \label{subsec:datasets}
% All datasets are open-source: PPMI \citep{marek2011parkinson}, IXI \citep{ixidata}, SALD \citep{wei2017structural}, and ADNI3/4 \citep{jack2008alzheimer}. Structural MRI and PET images were preprocessed using an FSL-based pipeline (reorientation, bias correction, registration to MNI). PET frames were averaged to static images. From each volume, the central coronal slice (and neighboring slices for generation tasks) was extracted. A complete description is in Appendix~\ref{app:preproc}.

\section{Results}
\label{sec:results}
In this section, we present the results of the experiments described in Section \ref{sec:experiments}. The generation results are reported in Section \ref{subsec:gen_results}, and the classification results are presented in Section \ref{subsec:class_results}.

\subsection{Image Generation Results}
\label{subsec:gen_results}

\begin{table}[ht]
\fontsize{8pt}{11pt}\selectfont
\renewcommand{\arraystretch}{1.5}
\caption{Comparison of generative models on scanner- and modality-conditioned tasks using FID variants. Bold indicates best scores; 95\% CIs are shown for FID$_\text{Rad}$ and KID$_\text{Rad}$.}
\vspace{1mm}
\centering
\label{tab:quality_metrics}
\begin{tabular}{>{\centering\arraybackslash}p{2.3cm}p{1.9cm}ccccc c c}
\toprule
\midrule
\textbf{Task} & \textbf{Model} & 
\textbf{FID} $\downarrow$ & 
\textbf{FID$_\text{Rad}$ $\downarrow$} & 
\textbf{KID$_\text{Rad}$ $\downarrow$} &
\textbf{FID$_\text{SwAV}$ $\downarrow$} &
\textbf{FID$^\text{b}_\text{Rad}$ (CI$_{\text{95\%}}$) $\downarrow$} &
\textbf{KID$^\text{b}_\text{Rad}$ (CI$_{\text{95\%}}$) $\downarrow$} \\
\midrule
\multirow{6}{1.2cm}{\centering \textbf{Scanner} \\ (\textit{Generation})} 
& SNGAN          &  29.91   &   3.03    &   2.4 \, \texttimes \,  10$^{\text{-3}}$    &    4.98    &  [2.95, 3.18]     &  [2.4,\ 2.7] \, \texttimes \, 10$^{\text{-3}}$     \\
& StyleGAN2  &  97.50   & 6.80  &  6.7 \, \texttimes \,  10$^{\text{-3}}$    &   9.33       & [6.64, 6.95]  &  [6.5,\ 6.9] \, \texttimes \, 10$^{\text{-3}}$    \\
& ADC-GAN    &  59.83      &  5.12     &    4.6 \, \texttimes \,  10$^{\text{-3}}$    &    5.21    &   [5.05, 5.31]   &    [4.5,\ 4.8] \, \texttimes \, 10$^{\text{-3}}$   \\
& DDPM           &  39.5   &    13.80   &   1.75 \, \texttimes \,  10$^{\text{-2}}$   &    5.74    &  [13.63, 14.01]     &   [1.73,\ 1.78] \, \texttimes \, 10$^{\text{-2}}$   \\
& CVAE  &  244.07  &   16.69    &       2.07 \, \texttimes \,  10$^{\text{-2}}$   &  21.56     &   [16.58,\  16.83]    &   [2.06,\ 2.10] \, \texttimes \, 10$^{\text{-2}}$     \\
& MAGIC-Flow   & \textbf{27.64}  & \textbf{0.84}   &   \textbf{6.7 \, \texttimes \,  10$^{\text{-4}}$ } &  \textbf{4.31}      &  \textbf{[0.83, 0.95]}    & \textbf{[6.2,\ 7.5] \, \texttimes \, 10$^{\text{-4}}$}    \\
\midrule
\multirow{6}{1.2cm}{\centering \textbf{Modality} \\ (\textit{Generation})} 
& SNGAN          &  34.06   &   3.82    &    2.6 \, \texttimes \,  10$^{\text{-3}}$    &  4.23      &   [3.76, 3.95]    &  [2.5,\ 2.7] \, \texttimes \, 10$^{\text{-3}}$     \\
& StyleGAN2    &  59.2   &  4.75   &    3.4 \, \texttimes \,  10$^{\text{-3}}$    &   6.50     & [4.68, 4.85]    &  [3.3,\ 3.5] \, \texttimes \, 10$^{\text{-3}}$   \\
& ADC-GAN        &  57.84   &   4.94    &    2.7 \, \texttimes \,  10$^{\text{-3}}$    &   5.46     &   [4.88, 5.07]    &  [2.4,\ 2.7] \, \texttimes \, 10$^{\text{-3}}$     \\
& DDPM           &   70.48  &   15.22    &   1.72 \, \texttimes \,  10$^{\text{-2}}$   &   8.08     &  [15.04,\  15.38]    &   [1.70,\ 1.74] \, \texttimes \, 10$^{\text{-2}}$     \\
& CVAE      &   215.88  &   14.71    &    1.47 \, \texttimes \,  10$^{\text{-2}}$   &  19.40      &  [14.56,\  14.88]    &   [1.45,\ 1.48] \, \texttimes \, 10$^{\text{-2}}$     \\
& MAGIC-Flow      &  \textbf{33.04}  & \textbf{0.98}   &   \textbf{4.5 \, \texttimes \,  10$^{\text{-4}}$} &  \textbf{4.07}      &  \textbf{[0.94, 1.08]}    & \textbf{[4.0,\ 5.2] \, \texttimes \, 10$^{\text{-4}}$}    \\     
\midrule
\bottomrule
\end{tabular}
\end{table}

% Real Dataset
% MS-SSIM intra: Mean: 0.6086, Std: 0.1154
% MS-SSIM inter: Mean: 0.2189, Std: 0.1848
\begin{table}[ht]
\fontsize{8pt}{11pt}\selectfont
\renewcommand{\arraystretch}{1.4}
\caption{Comparison of generative models using fidelity/diversity (P, R, D, C) and MS-SSIM. Gray highlights show the best-performing model; “Real Data” provides reference values.}
\vspace{1mm}
\centering
\label{tab:fid_and_div_metrics}
\begin{tabular}{>{\centering\arraybackslash}p{2.3cm}p{1.9cm}cccccc}
\toprule
\midrule
\textbf{Task} & \textbf{Model} & 
\textbf{P} $\uparrow$ & 
\textbf{R $\uparrow$} & 
\textbf{D $\uparrow$} & 
\textbf{C $\uparrow$} &
\textbf{MS-SSIM$^{intra}$ $\downarrow$} &
\textbf{MS-SSIM$^{inter}$ $\downarrow$} \\
\midrule
\multirow{6}{2.0cm}{\centering \textbf{Scanner} \\ (\textit{Generation})}
%----SNGAN
& SNGAN    &  0.79   &  0.05  &   0.77    &   0.40 &  0.68 $\pm$ 0.16   &   0.48 $\pm$ 0.09     \\
%----StyleGAN2
& StyleGAN2 &  0.99   &   0.0    &    0.61   &   0.04 &  0.99 $\pm$ 0.003   &   0.66 $\pm$ 0.10     \\
%----ADC-GAN
& ADC-GAN    &   0.94   &   0.01    &    0.72    &    0.10    & 0.89 $\pm$ 0.08   &   0.66 $\pm$ 0.08       \\
%----DDPM
& DDPM    &  0.0   &   1.0     &    0.0    &    0.0    &   0.46 $\pm$ 0.12   & 0.33 $\pm$ 0.12     \\
%----CVAE
& CVAE   &  0.0   &    0.0    &   0.0    &   0.0     &  0.99 $\pm$ 0.002  & 0.83 $\pm$ 0.06     \\
%----MAGIC-Flow
& \cellcolor{gray!20} MAGIC-Flow  
& \cellcolor{gray!20} 0.87  
& \cellcolor{gray!20} 0.64    
& \cellcolor{gray!20} 0.91    
& \cellcolor{gray!20} 0.84    
& \cellcolor{gray!20} 0.60 $\pm$ 0.08 
& \cellcolor{gray!20} 0.49 $\pm$ 0.09 \\
& Real Data  &  -  &   -   &   -   &   -    &   0.51 $\pm$ 0.11 & 0.43 $\pm$ 0.11        \\
\midrule
\multirow{6}{2.0cm}{\centering \textbf{Modality} \\ (\textit{Generation})} 
%----SNGAN
& SNGAN    &  0.89  &  0.0  & 0.98 & 0.33  &   0.79 $\pm$ 0.18 & 0.22 $\pm$ 0.19 \\
%----StyleGAN2
& StyleGAN2 &  0.98 & 0.02 & 1.0 & 0.25 &  0.90 $\pm$ 0.07 & 0.29 $\pm$ 0.25  \\
%----ADC-GAN
& ADC-GAN    &   1.0   &   0.0     &   0.87     &    0.17    &  0.87 $\pm$ 0.11   &   0.23 $\pm$ 0.21       \\
%----DDPM
& DDPM    &  0.0  &   0.0   &    0.0    &   0.0     &  0.55 $\pm$ 0.12   &   0.20 $\pm$ 0.18      \\
%----CVAE
& CVAE   &  0.0  &    0.0    &    0.0    &    0.0    &  0.99 $\pm$ 0.003   &   0.32 $\pm$ 0.29     \\
%----MAGIC-Flow 
& \cellcolor{gray!20} MAGIC-Flow
& \cellcolor{gray!20} 0.94  
& \cellcolor{gray!20} 0.64    
& \cellcolor{gray!20} 1.0    
& \cellcolor{gray!20} 0.87    
& \cellcolor{gray!20} 0.64 $\pm$ 0.09 
& \cellcolor{gray!20} 0.22 $\pm$ 0.18 \\
& Real Data  &  -  &   -   &   -   &   -    &   0.61 $\pm$ 0.12 & 0.22 $\pm$ 0.18        \\
\midrule
\bottomrule
\end{tabular}
\end{table}

\begin{figure}[ht]
\centering
\begin{overpic}[width=0.95\linewidth]{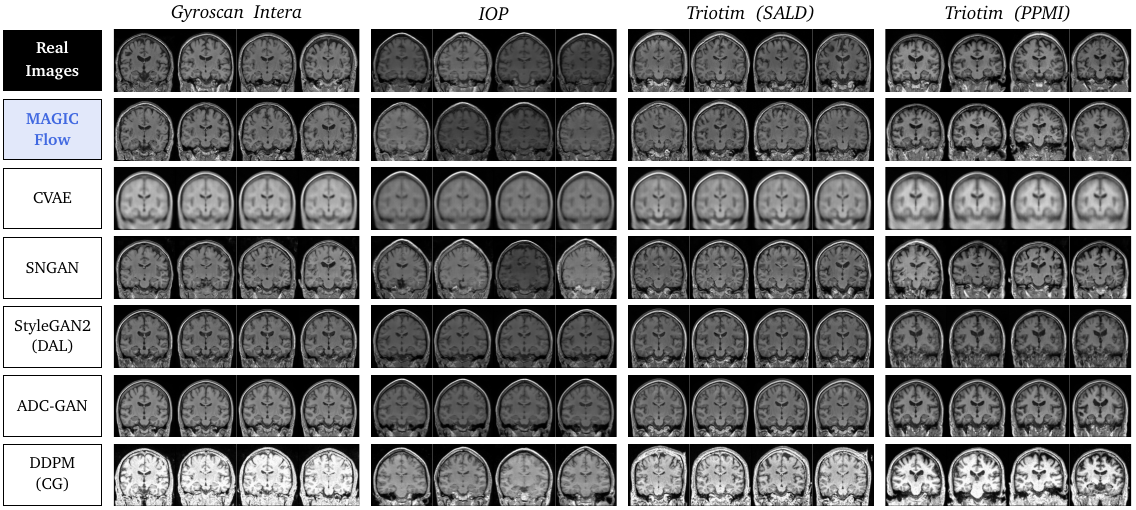}
    \put(102,20){\small (a)} 
\end{overpic}

\vspace{0.2cm}

\begin{overpic}[width=0.95\linewidth]{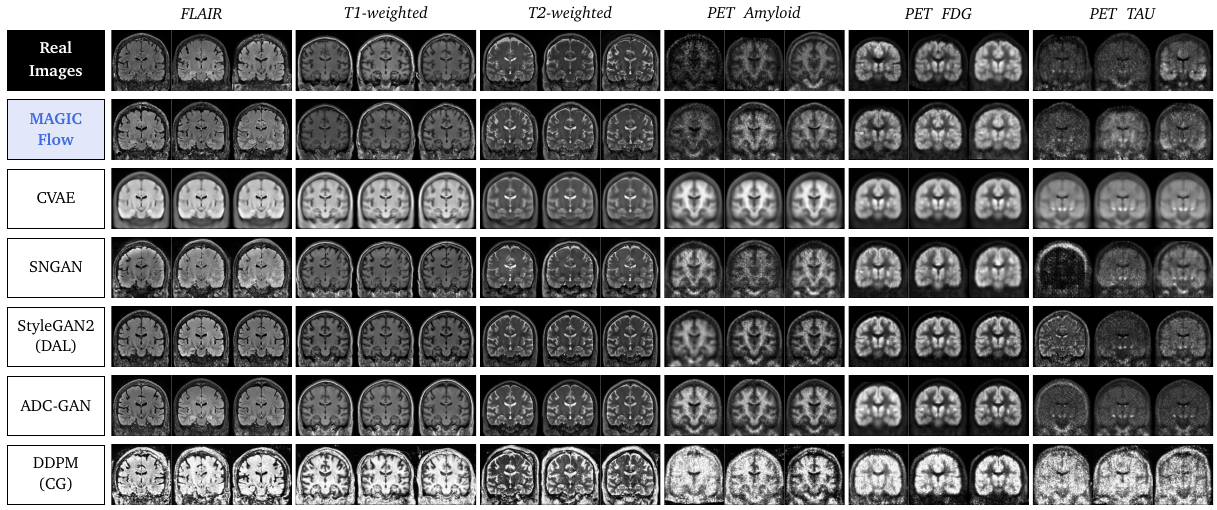}
    \put(102,20){\small (b)} 
\end{overpic}

\caption{Qualitative comparison of generated neuroimaging slices across (a) scanner types and (b) imaging modalities.}
\label{fig:montage_combined}
\end{figure}

Quantitative results are summarized in Tables~\ref{tab:quality_metrics} and \ref{tab:fid_and_div_metrics}.
MAGIC-Flow consistently outperforms all baselines across both tasks. For scanner-conditioned generation, it achieves an FID$_\text{Rad}$ of 0.84 and KID$_\text{Rad}$ of $6.7 \times 10^{-4}$, substantially lower than the closest baseline (SNGAN: 3.03 and $2.4 \times 10^{-3}$). Similar trends are observed for modality-conditioned generation (FID$_\text{Rad}$ 0.98, KID$_\text{Rad}$ $4.5 \times 10^{-4}$), with bootstrapped confidence intervals confirming robust performance. Unlike GAN-based baselines, which often trade fidelity for diversity (e.g., SNGAN: P=0.79, R=0.05, D = 0.77, C = 0.40), MAGIC-Flow achieves balanced performance with high precision (0.87) and substantially higher recall (0.64), along with the highest density (0.91) and coverage (0.84), indicating realistic and representative samples. MS-SSIM scores further show that intra-class variability closely matches real data, while inter-class separability is maintained.

Qualitative comparisons, shown in Figure~\ref{fig:montage_combined}, confirm these findings: MAGIC-Flow generates sharp, anatomically consistent images preserving scanner- and modality-specific features, whereas CVAE produces overly smooth outputs, GANs collapse to repetitive patterns, and DDPMs, though diverse, often appear noisy or implausible.

In terms of efficiency, MAGIC-Flow generates 37.5 images/s, offering a favorable trade-off between quality and speed compared to other approaches (SNGAN: 49.75; ADC-GAN: 38.17; StyleGAN2: 52.20; CVAE: 369.0; DDPM: 0.24).

\subsection{Scanner Classification}
\label{subsec:class_results}

\begin{table}[ht]
\fontsize{8pt}{11pt}\selectfont
\renewcommand{\arraystretch}{1.5}
\caption{Comparison of classification performance on the test set.}
\vspace{1mm}
\centering
\label{tab:classification_metrics}
\begin{tabular}{m{1.9cm} c c c c c c c}
\toprule
\midrule
\textbf{Model} & \textbf{Pre-Trained} & \textbf{Accuracy} & \textbf{Balanced Acc.} & \textbf{AUC} & \textbf{Precision} & \textbf{Recall} & \textbf{F1-score} \\
\midrule
MAGIC-Flow & - & 0.90 $\pm$ 0.01 & 0.76 $\pm$ 0.02 & 0.97 $\pm$ 0.01 & 0.77 $\pm$ 0.03 & 0.76 $\pm$ 0.03 & 0.75 $\pm$ 0.03 \\
ResNet-50 & RadImageNet & 0.91 $\pm$ 0.01 & 0.73 $\pm$ 0.02 & 0.98 $\pm$ 0.002 & 0.71 $\pm$ 0.06 & 0.73 $\pm$ 0.02 & 0.71 $\pm$ 0.04 \\
DenseNet-121 & RadImageNet & 0.91 $\pm$ 0.01 & 0.73 $\pm$ 0.03 & 0.98 $\pm$ 0.00 & 0.79 $\pm$ 0.03 & 0.73 $\pm$ 0.03 & 0.72 $\pm$ 0.03 \\
InceptionV3 & RadImageNet & 0.90 $\pm$ 0.02 & 0.72 $\pm$ 0.02 & 0.97 $\pm$ 0.003 & 0.67 $\pm$ 0.04 & 0.72 $\pm$ 0.02 & 0.69 $\pm$ 0.03 \\
IncepResNetV2 & RadImageNet & 0.88 $\pm$ 0.02 & 0.74 $\pm$ 0.03 & 0.97 $\pm$ 0.01 & 0.76 $\pm$ 0.03 & 0.74 $\pm$ 0.03 & 0.74 $\pm$ 0.03 \\  
ViT & - & 0.88 $\pm$ 0.01 & 0.70 $\pm$ 0.02 & 0.97 $\pm$ 0.003 & 0.72 $\pm$ 0.02 & 0.70 $\pm$ 0.02 & 0.70 $\pm$ 0.02 \\
ViT-ResNet & ImageNet-21k & 0.91 $\pm$ 0.01 & 0.73 $\pm$ 0.03 & 0.98 $\pm$ 0.00 & 0.74 $\pm$ 0.09 & 0.73 $\pm$ 0.03 & 0.71 $\pm$ 0.04  \\
Swin-ViT & ImageNet-22k & 0.91 $\pm$ 0.004 & 0.73 $\pm$ 0.03 & 0.98 $\pm$ 0.01 & 0.71 $\pm$ 0.07 & 0.73 $\pm$ 0.03 & 0.71 $\pm$ 0.05 \\
\midrule
\bottomrule
\end{tabular}
\end{table}

\begin{figure}[ht]
\begin{center}
\includegraphics[width=1.0\textwidth]{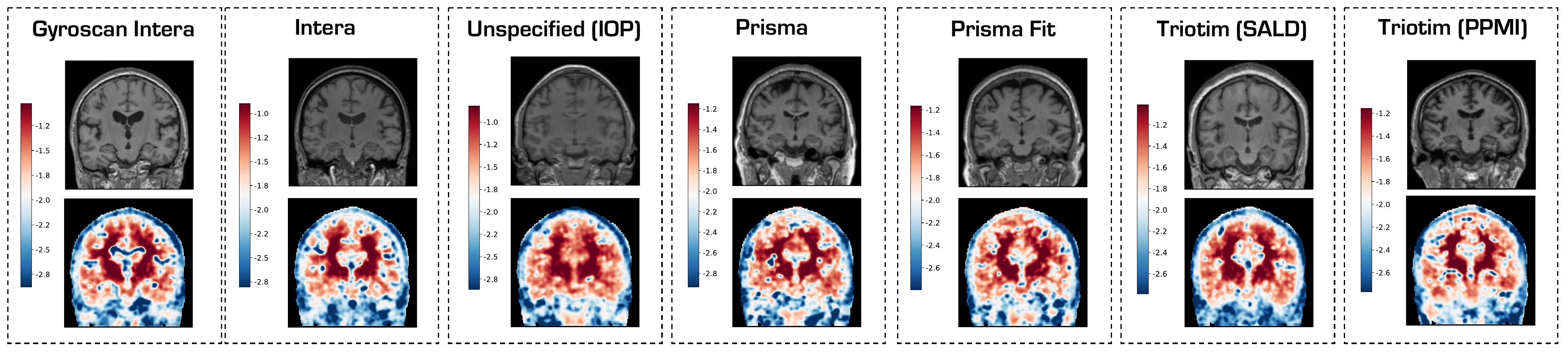}
\end{center}
\caption{Likelihood attribution maps from MAGIC-Flow. Warm colors indicate regions with the strongest positive contributions to the predicted likelihood.}
\label{fig:heatmaps}
\end{figure}

As shown in Table~\ref{tab:classification_metrics}, MAGIC-Flow achieves $0.90 \pm 0.01$ accuracy, comparable to CNN and ViT baselines, while surpassing them in balanced accuracy ($0.76 \pm 0.02$), macro recall ($0.76 \pm 0.03$), and F1-score ($0.75 \pm 0.03$), demonstrating enhanced robustness to underrepresented scanners and discrimination of \textit{close} classes. MAGIC-Flow is shown to combine competitive predictive performance with improved fairness across classes and provides interpretable, likelihood-based explanations.

Figure~\ref{fig:heatmaps} shows likelihood-based attribution maps.  MAGIC-Flow consistently attends to global intracranial intensity distributions, particularly along gray–white matter boundaries, around ventricles, and across cortical regions, while remaining insensitive to subject anatomy. This confirms the model relies on acquisition signatures rather than anatomical variability.

\section{Conclusion} 
\label{sec:discussion}

We introduced MAGIC-Flow, a conditional multiscale normalizing flow that unifies generation and classification within a likelihood-based framework. It enables high-fidelity synthesis, robust classification under scanner variability and imbalance, and interpretable likelihood attribution maps. Current limitations include its 2D scope; extending to 3D, adding uncertainty measures, and exploring pathology-conditioned generation and multi-institutional deployment are future directions. A key use is augmenting scarce datasets of underrepresented conditions. MAGIC-Flow advances joint generative-discriminative modeling for data-limited, privacy-sensitive domains.

\clearpage
\phantomsection

\section*{Acknowledgements}
L. Caldera is funded by Health Big Data project sponsored by the Italian Ministry of Health (CCR-2018-23669122). L. Cavinato is funded by the National Plan for NRRP Complementary Investments “Advanced Technologies for Human-centred Medicine” (PNC0000003). Giacomo Bottacini acknowledges the support of European Union - NextGenerationEU within the Italian PNRR program (M4C2, Investment 3.3) for the PhD Scholarship ‘‘Physics-informed Artificial Intelligence for surrogate modeling of complex energy systems’’. The present research is part of the activities of “Dipartimento di Eccellenza 2023-2027".

Data collection and sharing for this project was funded by the Alzheimer's Disease Neuroimaging Initiative (ADNI) (National Institutes of Health Grant U01 AG024904) and DOD ADNI (Department of Defense award number W81XWH-12-2-0012). ADNI is funded by the National Institute on Aging, the National Institute of Biomedical Imaging and Bioengineering, and through generous contributions from the following: AbbVie, Alzheimer’s
Association; Alzheimer’s Drug Discovery Foundation; Araclon Biotech; BioClinica, Inc.; Biogen; Bristol-Myers Squibb Company; CereSpir, Inc.; Cogstate; Eisai Inc.; Elan Pharmaceuticals, Inc.; Eli Lilly and Company; EuroImmun; F. Hoffmann-La Roche Ltd and its affiliated company Genentech, Inc.; Fujirebio; GE Healthcare; IXICO Ltd.; Janssen Alzheimer Immunotherapy Research \& Development, LLC.; Johnson \& Johnson
Pharmaceutical Research \& Development LLC.; Lumosity; Lundbeck; Merck \& Co., Inc.; Meso Scale Diagnostics, LLC.; NeuroRx Research; Neurotrack Technologies; Novartis
Pharmaceuticals Corporation; Pfizer Inc.; Piramal Imaging; Servier; Takeda Pharmaceutical Company; and Transition Therapeutics. The Canadian Institutes of Health Research is
providing funds to support ADNI clinical sites in Canada. Private sector contributions are facilitated by the Foundation for the National Institutes of Health (www.fnih.org). The grantee organization is the Northern California Institute for Research and Education, and the study is coordinated by the Alzheimer’s Therapeutic Research Institute at the University of Southern California. ADNI data are disseminated by the Laboratory for Neuro Imaging at the University of Southern California.

\bibliographystyle{unsrt}  
%\bibliography{library}  %%% Remove comment to use the external .bib file (using bibtex).


\newpage
\appendix

\section*{\textbf{SUPPLEMENTARY MATERIAL}}

\section{Mathematical and Architectural Details}
\label{app:math}

\subsection{Conditional Change of Variables Formula: Detailed Derivation}
\label{app:conditional-cov}

Let $Z \in \mathbb{R}^d$ and $Y \in \mathbb{R}^k$ be random variables with joint density $p_{Z,Y}(z,y)$.  
Let
\[
f : \mathbb{R}^d \times \mathbb{R}^k \to \mathbb{R}^d
\]
be continuously differentiable, and suppose that for every fixed $y$, the map $z \mapsto f(z,y)$ is a diffeomorphism onto its image with Jacobian matrix $\tfrac{\partial f}{\partial z}(z,y)$ having nonzero determinant.  
Define $X := f(Z,Y)$, and write $f^{-1}(x,y)$ for the unique $z$ such that $f(z,y)=x$.  
Assume $p_Y(y) > 0$.  

If $Z$ and $Y$ are independent, the conditional density of $X$ given $Y=y$ is
\[
p_{X \mid Y}(x \mid y)
= p_Z\big(f^{-1}(x,y)\big) \;
\Big|\det\!\Big(\frac{\partial f^{-1}}{\partial x}(x,y)\Big)\Big|.
\]

\paragraph{Derivation.} Consider the mapping
\[
(Z, Y) \mapsto (X, Y) = (f(Z, Y), Y),
\]
with Jacobian
\[
\frac{\partial (x, y)}{\partial (z, y)} =
\begin{bmatrix}
\frac{\partial f(z, y)}{\partial z} & \frac{\partial f(z, y)}{\partial y} \\
0 & I_k
\end{bmatrix},
\]
where $I_k$ is the $k \times k$ identity matrix.  
Since this is block lower-triangular, its determinant is
\[
\left| \det \left( \frac{\partial (x, y)}{\partial (z, y)} \right) \right|
= \left| \det \left( \frac{\partial f(z, y)}{\partial z} \right) \right|.
\]

By the change-of-variables formula, the joint density of $(X, Y)$ is
\[
p_{X, Y}(x, y) = p_{Z, Y}(f^{-1}(x, y), y) \cdot 
\left| \det \left( \frac{\partial f}{\partial z}(f^{-1}(x, y), y) \right) \right|^{-1}.
\]

If $Z$ and $Y$ are independent, then
\[
p_{Z, Y}(z, y) = p_Z(z) \cdot p_Y(y).
\]

Substituting, we obtain
\[
p_{X, Y}(x, y) = p_Z(f^{-1}(x, y)) \cdot p_Y(y) \cdot 
\left| \det \left( \frac{\partial f}{\partial z}(f^{-1}(x, y), y) \right) \right|^{-1}.
\]

The conditional density follows from
\[
p_{X \mid Y}(x \mid y) = \frac{p_{X, Y}(x, y)}{p_Y(y)}.
\]

Using the Jacobian determinant of the inverse function,
\[
\left| \det \left( \frac{\partial f}{\partial z}(f^{-1}(x, y), y) \right) \right|^{-1} 
= \left| \det \left( \frac{\partial f^{-1}}{\partial x}(x, y) \right) \right|,
\]
we obtain the final expression:
\[
p_{X \mid Y}(x \mid y) = p_Z(f^{-1}(x, y)) \cdot 
\left| \det \left( \frac{\partial f^{-1}}{\partial x}(x, y) \right) \right|.
\]

This result provides the conditional version of the change-of-variables formula used in our model.

\subsection{Affine Coupling: Jacobian Derivation}
\label{app:inv_transf}

Let $\mathbf{x} \in \mathbb{R}^{C\times H\times W}$ be the input image and let $\mathbf{M}\in\{0,1\}^{C\times H\times W}$ be a binary mask. Define the complementary parts
\[
\mathbf{x}_A = \mathbf{M} \odot \mathbf{x}, \qquad
\mathbf{x}_B = (\mathbf{1}-\mathbf{M}) \odot \mathbf{x}.
\]

\paragraph{First affine transformation (A $\to$ B).}
The update equations are
\[
\mathbf{u}_A = \mathbf{x}_A, \qquad
\mathbf{u}_B = \mathbf{x}_B \odot \exp\big(\mathbf{s}_1(\mathbf{x}_A;\mathbf{y})\big) + \mathbf{t}_1(\mathbf{x}_A;\mathbf{y}),
\]
where $\mathbf{s}_1, \mathbf{t}_1 \in \mathbb{R}^{C\times H\times W}$ are elementwise scale and translation factors.  

The Jacobian is
\[
J_1 = \frac{\partial (\mathbf{u}_A, \mathbf{u}_B)}{\partial (\mathbf{x}_A, \mathbf{x}_B)}.
\]

Since $\mathbf{u}_A = \mathbf{x}_A$, we have
\[
\frac{\partial \mathbf{u}_A}{\partial \mathbf{x}_A} = I, \qquad 
\frac{\partial \mathbf{u}_A}{\partial \mathbf{x}_B} = 0.
\]

Each element of $\mathbf{u}_B$ depends elementwise on $x_{B,i}$ via
\[
u_{B,i} = x_{B,i} \cdot \exp(s_{1,i}(\mathbf{x}_A;\mathbf{y})) + t_{1,i}(\mathbf{x}_A;\mathbf{y}),
\]
so
\[
\frac{\partial u_{B,i}}{\partial x_{B,i}} = \exp(s_{1,i}(\mathbf{x}_A;\mathbf{y})).
\]

The Jacobian matrix is thus block-triangular:
\[
J_1 = 
\begin{bmatrix}
\frac{\partial \mathbf{u}_A}{\partial \mathbf{x}_A} & \frac{\partial \mathbf{u}_A}{\partial \mathbf{x}_B} \\
\frac{\partial \mathbf{u}_B}{\partial \mathbf{x}_A} & \frac{\partial \mathbf{u}_B}{\partial \mathbf{x}_B}
\end{bmatrix} =
\begin{bmatrix}
I & 0 \\
* & \operatorname{diag}\big(\exp(\mathbf{s}_1)\big)
\end{bmatrix},
\]
where $*$ represents entries that do not affect the determinant. Therefore, the log-determinant of the first transformation is
\[
\log|\det J_1| = \sum_{c,h,w} (1-M_{c,h,w}) \, s_{1,c,h,w}.
\]

\paragraph{Second affine transformation (B $\to$ A).}
The second transformation updates $\mathbf{u}_A$ while keeping $\mathbf{u}_B$ fixed:
\[
\mathbf{x}'_B = \mathbf{u}_B, \qquad
\mathbf{x}'_A = \mathbf{u}_A \odot \exp\big(\mathbf{s}_2(\mathbf{u}_B;\mathbf{y})\big) + \mathbf{t}_2(\mathbf{u}_B;\mathbf{y}).
\]

Again, the Jacobian is block-triangular, with
\[
\frac{\partial \mathbf{x}'_A}{\partial \mathbf{u}_A} = \operatorname{diag} \big(\exp(\mathbf{s}_2)\big),
\qquad
\frac{\partial \mathbf{x}'_B}{\partial \mathbf{u}_B} = I.
\]

Thus the log-determinant of the second transformation is
\[
\log|\det J_2| = \sum_{c,h,w} M_{c,h,w} \, s_{2,c,h,w}.
\]

\paragraph{Total log-determinant.}
The two transformations are composed sequentially, so the total log-determinant is
\[
\log|\det J| = \log|\det J_1| + \log|\det J_2| 
= \sum_{c,h,w} (1-M_{c,h,w})\, s_{1,c,h,w} + \sum_{c,h,w} M_{c,h,w}\, s_{2,c,h,w}.
\]

\subsection{Design of Affine Coupling Layers}
\label{app:aff_cou_arch}

\begin{figure}[ht]
\centering
\begin{subfigure}{0.49\textwidth}
    \centering
    \includegraphics[width=0.85\linewidth]{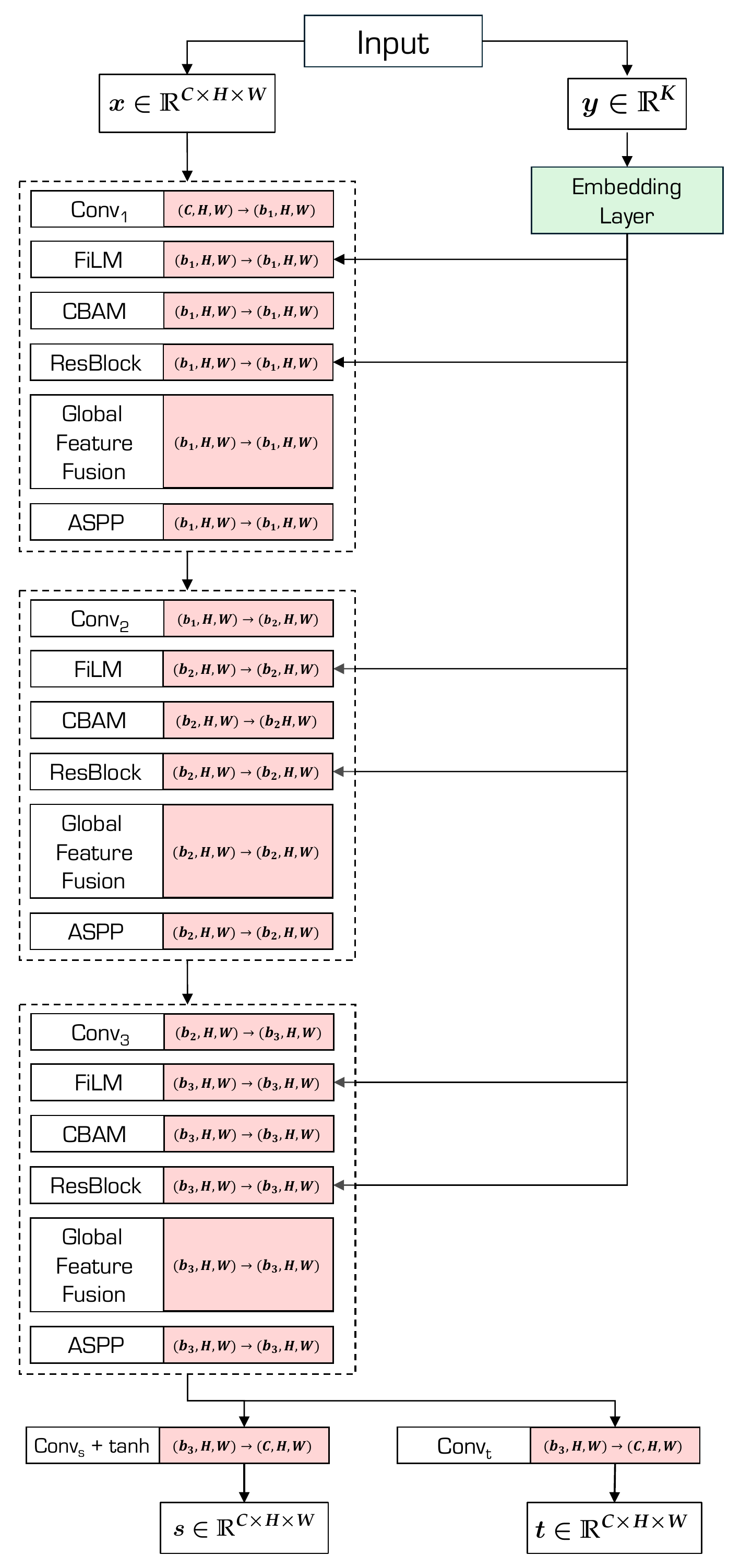}
    \caption{Generation coupling layer: designed for expressive, multi-scale, and context-aware transformations to synthesize high-fidelity outputs.}
\label{fig:gen_aff_cou}
\end{subfigure}
\hfill
\begin{subfigure}{0.49\textwidth}
    \centering
    \includegraphics[width=\linewidth]{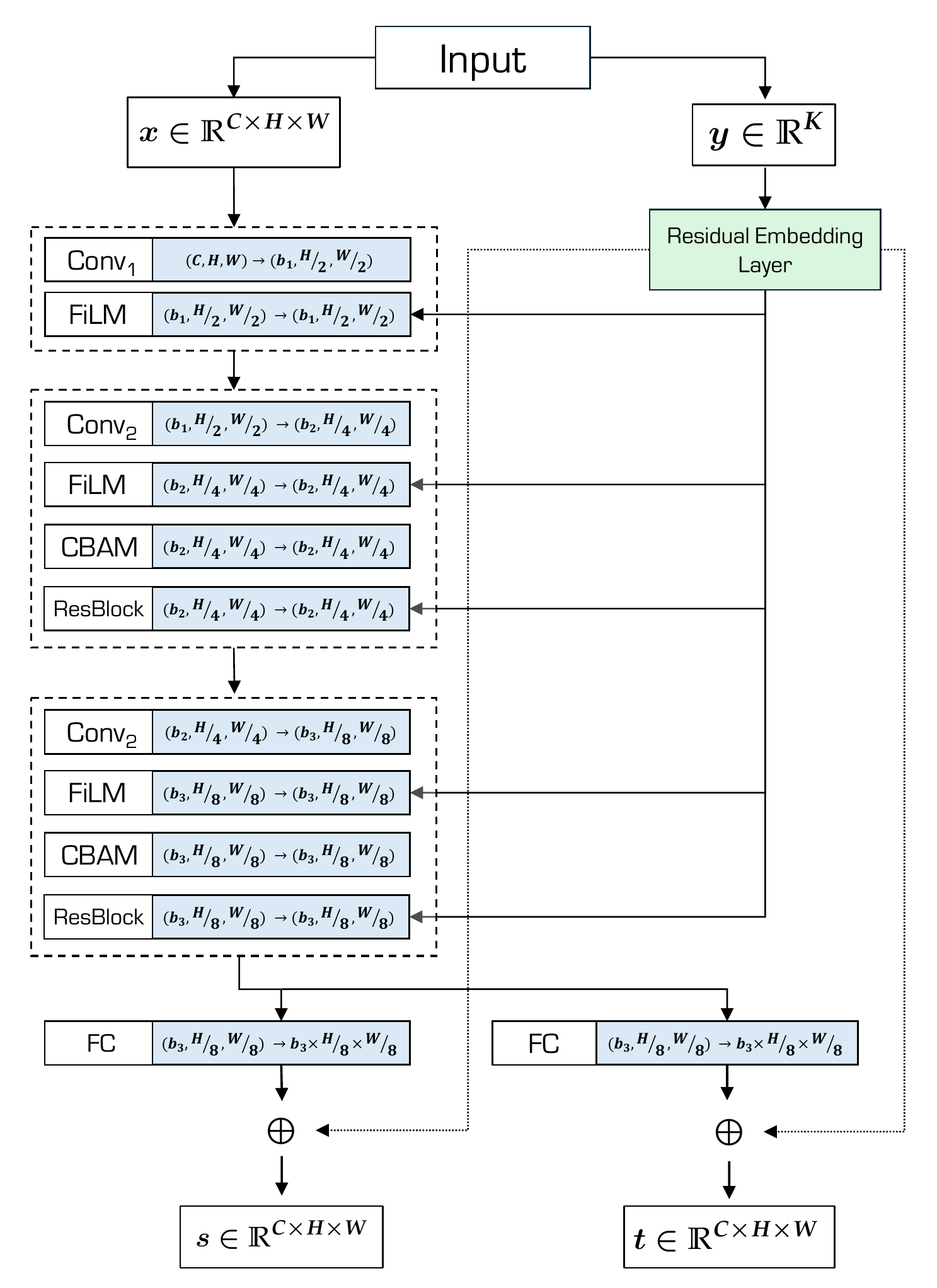}
    \caption{Classification coupling layer: optimized for label-aware, discriminative feature transformations for predictive accuracy.}
    \label{fig:class_aff_cou}
\end{subfigure}
\caption{Architectures of the task-specific affine coupling layers. The generation layer emphasizes multi-scale and global context features, while the classification layer focuses on integrating label information into the feature transformations.}
\label{fig:aff_cou_architecture}
\end{figure}

Figure~\ref{fig:aff_cou_architecture} illustrates the task-specific affine coupling layers used in MAGIC-Flow. The generation-oriented coupling layer (Figure~\ref{fig:gen_aff_cou}) is designed to capture multi-scale dependencies and global context, enabling expressive transformations for high-fidelity sample synthesis. In contrast, the classification-oriented coupling layer (Figure~\ref{fig:class_aff_cou}) incorporates label information directly into the feature transformations, prioritizing discriminative structure to improve predictive accuracy. Together, these designs highlight the flexibility of our framework in tailoring flow-based transformations to the demands of different tasks.

\subsection{Mask Design} 
\label{app:mask_design}

To ensure that every dimension of $\mathbf{x}$ is transformed throughout the flow, MAGIC-Flow alternates among three types of binary masks $\mathbf{M} \in \{0,1\}^{C \times H \times W}$:  
(1) \textbf{Checkerboard masks:} update alternating spatial positions to encourage pixel-level mixing;  
(2) \textbf{Channel-wise masks:} transform subsets of channels, enabling feature-level diversity;  
(3) \textbf{Application-specific masks:} emphasize semantically relevant regions, adapting transformations to the downstream task.  

By alternating checkerboard and channel-wise masks across flow steps, MAGIC-Flow achieves a balance between spatial and feature-level expressiveness while preserving tractable Jacobian computations.  

\begin{figure}[ht]
\begin{center}
\includegraphics[width=1.0\textwidth]{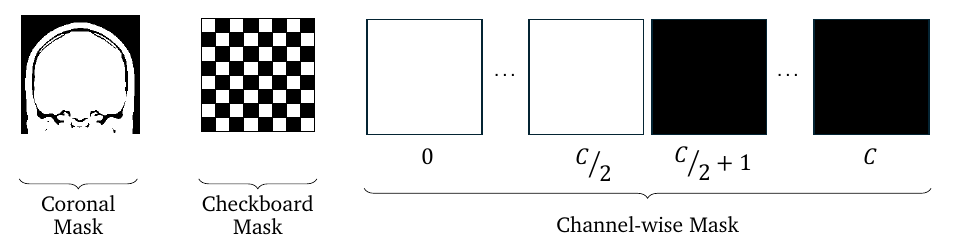}
\end{center}
\caption{Examples of the masks used in MAGIC-Flow: from left to right, (1) application-specific coronal mask, (2) checkerboard mask, and (3) channel-wise mask. Together, these masks provide complementary spatial, feature-level, and application-adaptive transformations.}
\label{fig:masks}
\end{figure}

The masks used in our experiments are shown in Figure~\ref{fig:masks}. Because our applications involve coronal slices, the application-specific mask is designed as a coronal binary mask. We find that incorporating this application-specific mask reduces artifacts in the generated images.  

\subsection{Likelihood Attribute Maps}
\label{app:likelihood}
\begin{figure}[ht]
\begin{center}
\includegraphics[width=1.0\textwidth]{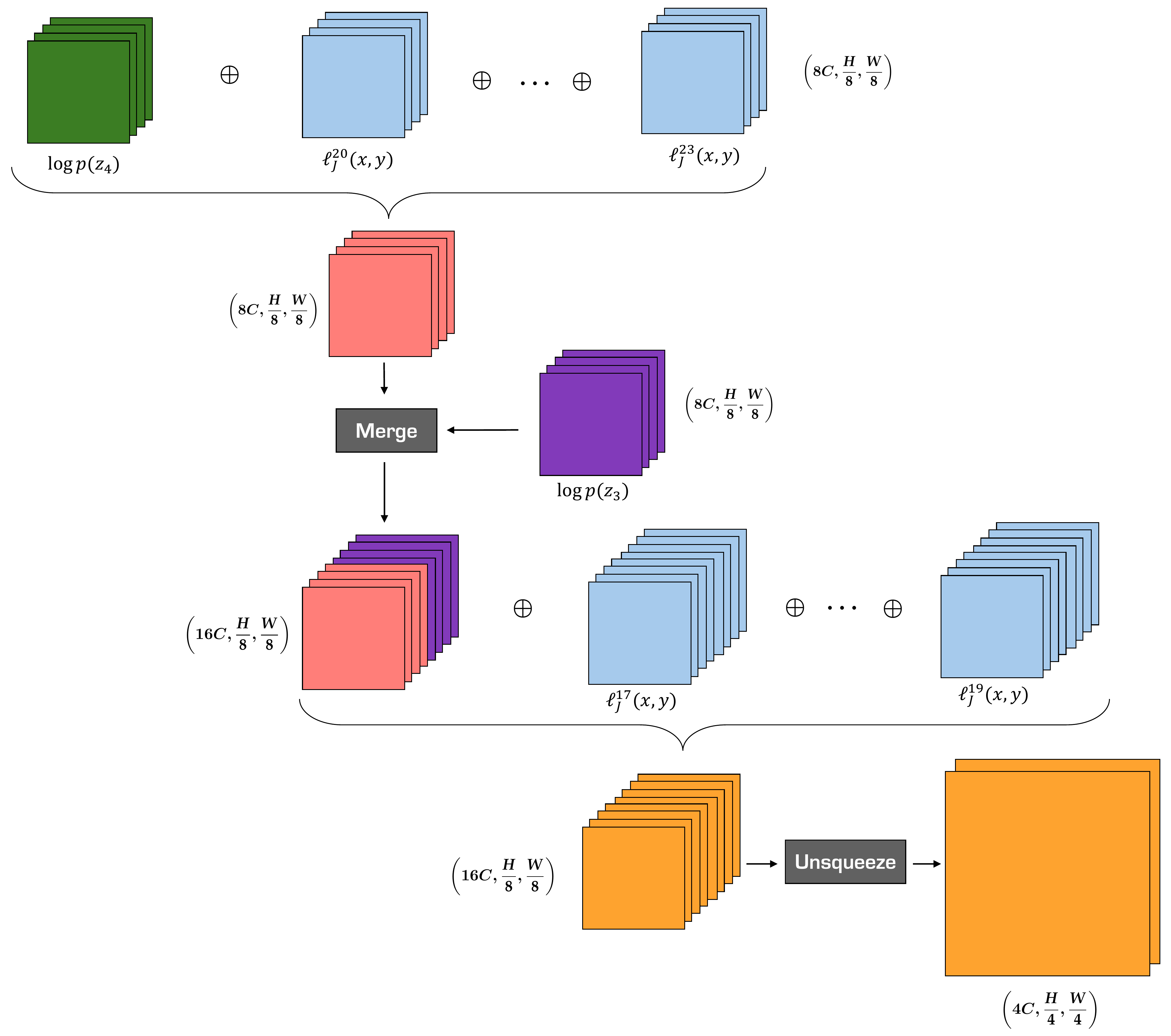}
\end{center}
\caption{Illustration of the initial steps in constructing the maps described in Section \ref{sec:heatmaps}. Starting from latent factors (green) at the deepest level, contributions are propagated backward through inverse split and squeeze operations, with the corresponding Jacobian terms accumulated at each step. This process produces a pixel-level attribution map $\mathcal{H}(\mathbf{x}, \mathbf{y})$, providing a principled, likelihood-based interpretation of MAGIC-Flow’s predictions.}
\label{fig:deriv_heatmaps}
\end{figure}

To better understand the predictive behavior of MAGIC-Flow, we introduce
\emph{Likelihood Attribute Maps}. These maps provide a pixel-level attribution
that links the model’s latent likelihood computations to the input domain.
As illustrated in Figure~\ref{fig:deriv_heatmaps}, the method propagates
contributions from deep latent factors back to the input space through inverse
split and squeeze operations, while accounting for the Jacobian terms at each
stage. The resulting attribution map $\mathcal{H}(\mathbf{x}, \mathbf{y})$
offers an interpretable, likelihood-grounded explanation of the model’s output.

\section{Supplementary Experimental Details}
\label{app:experiments}

\subsection{Imaging Preprocessing Pipeline}
\label{app:preproc}
All imaging data, including structural MRI and PET scans, were preprocessed using a customized pipeline based on the FMRIB Software Library (FSL) \cite{fsl1, fsl2}. T1-weighted images were processed with FSL’s \texttt{fsl\_anat} pipeline, which included reorientation to standard space, bias-field correction and nonlinear registration to the MNI152 template. T2-weighted and FLAIR images were rigidly coregistered to the subject’s T1-weighted image using normalized mutual information cost functions, with or without bias-field correction depending on data quality. The resulting transformations were applied to align T2-weighted and FLAIR scans to T1 space and subsequently to MNI space using the T1-derived warp fields.
All PET data were obtained from Alzheimer's Disease Neuroimaging Initiative (ADNI3 and ADNI4). FDG, amyloid, and tau PET scans were first coregistered to the subject’s T1-weighted image and then normalized to MNI space using the T1-derived transformations. In ADNI, PET acquisitions occur during the plateau phase of tracer uptake (e.g., 30–60 min post-injection for FDG, 50–70 min for florbetapir and NAV-4694, 90–110 min for florbetaben and MK-6240, 75–105 min for flortaucipir, and 45–75 min for PI-2620) and are reconstructed into multiple short frames. Specifically, florbetapir, florbetaben, and NAV-4694 scans are acquired as 4 × 5 min frames, whereas FDG, flortaucipir, and PI-2620 scans are acquired as 6 × 5 min frames. To generate a single static image representing tracer distribution, the median across frames was retained. The median was selected over the mean because it provides a more robust estimate of plateau-phase uptake, minimizing the influence of motion artifacts or frame-specific outliers that could bias the averaged signal. This multimodal preprocessing pipeline ensured consistent alignment of structural (T1, T2, FLAIR) and molecular (FDG, amyloid, tau PET) imaging data across participants.

After preprocessing, the resulting images had dimensions of $(1, \ 182, \ 218, \ 182)$. For our experiments, each 3D volume was processed by extracting coronal slices of shape $(1, \ 182, \ 182)$. Specifically, we selected the central coronal slice from each volume. To augment the datasets while maintaining anatomical consistency, neighboring slices were also included only for the scanner-conditioned and modality-conditioned generation tasks, typically within ±5 slices of the center. In some cases, fewer slices were used depending on the dataset size. No slice augmentation was applied for the scanner classification task.

\subsection{Imaging Datasets Overview}
\label{app:datasets}

\begin{table}[ht]
\fontsize{8pt}{11pt}\selectfont
\renewcommand{\arraystretch}{1.5}
\centering
\caption{Overview of imaging datasets used across the three tasks.
(a) Scanner-conditioned generation and scanner classification: scanner models and manufacturers, along with the number of coronal slices available for each scanner in both classification and generation tasks.
(b) Modality-conditioned generation: MRI sequences and PET tracers, with the corresponding number of coronal slices per modality and total counts.}
\vspace{1mm}
% ----------------- Table (a) -----------------
\begin{subtable}[t]{1\textwidth} % adjust width as needed
\centering
\begin{tabular}{p{2.2cm} p{2.0cm} c c c}
\toprule
\textbf{Scanner Model} & \textbf{Manufacturer} & \textbf{Dataset} & \textbf{\# Coronal slices (Classification)} & \textbf{\# Coronal slices (Generation)} \\
\midrule
Gyroscan Intera        & Philips  & IXI   & 322 & 349 \\
Intera                 & Philips  & IXI   & 185 & -- \\
Unspecified (IOP)      & GE       & IXI   & 74  & 310 \\
TrioTim                & SIEMENS  & PPMI  & 41  & 315 \\
TrioTim                & SIEMENS  & SALD  & 494 & 345 \\
Prisma                 & SIEMENS  & ADNI3 & 69  & -- \\
Prisma Fit             & SIEMENS  & ADNI3 & 167 & -- \\
\bottomrule
\end{tabular}
%\vspace{0.1cm}
\caption{}
\end{subtable}
\hfill
% ----------------- Table (b) -----------------
\begin{subtable}[t]{1\textwidth} % adjust width as needed
\centering
\label{tab:table_b}
\begin{tabular}{c c c c c}
\toprule
\textbf{Modality} & \textbf{Sequence} & \textbf{Tracers} & \textbf{Dataset} & \textbf{\# Coronal slices} \\
\midrule
MRI & FLAIR        & N/A           & ADNI3 & 526  \\
MRI & T1-weighted  & N/A           & ADNI3 & 527  \\
MRI & T2-weighted  & N/A           & ADNI4 & 535  \\
PET & Amyloid      & AV45 - FBB - NAV4694 & ADNI3 / ADNI4 & 516  \\
PET & Tau          & AV1451 - MK6240 - PI2620 & ADNI3 / ADNI4  & 518  \\
PET & FDG          & 18F-FDG       & ADNI3 & 522  \\  
\bottomrule
\end{tabular}
%\vspace{0.1cm}
\caption{}
\end{subtable}
\label{tab:datasets}
\end{table}

\subsubsection{Datasets for Generation}

Two conditional generation tasks were evaluated: (i) \textit{scanner-conditioned generation}, assessing whether MAGIC-Flow can synthesize realistic MRI slices that preserve scanner-specific characteristics, and (ii) \textit{modality-conditioned generation}, testing the model’s ability to generate anatomically consistent images across MRI (T1, T2, FLAIR) and PET (FDG, amyloid, tau) modalities.
 
Scanner-conditioned experiments used T1-weighted MRI scans from PPMI \citep{marek2011parkinson}, IXI \citep{ixidata}, and SALD \citep{wei2017structural}, with the following slice counts: PPMI (TrioTim, 41 slices for classification, 315 for generation), IXI (Gyroscan Intera: 322/349, Intera: 185/–, GE unspecified: 74/310), and SALD (TrioTim: 494/345).

Modality-conditioned experiments used multimodal MRI and PET data from ADNI3/4 \citep{jack2008alzheimer}. Central coronal slices along with ±5 neighboring slices were extracted from MRI (FLAIR, T1, T2) and PET (FDG, amyloid, tau) scans. Slice counts per class were: MRI-FLAIR (526), MRI-T1 (527), MRI-T2 (535), PET-Amyloid (516), PET-Tau (518), and PET-FDG (522), yielding a total of 3,144 slices across all modalities.

\subsubsection{Datasets for Classification}

Scanner classification evaluates MAGIC-Flow’s ability to discriminate among seven scanner models, including closely related models (e.g., Siemens Prisma vs. Prisma Fit), using unbalanced datasets to test robustness.
Coronal slices from PPMI, SALD, IXI, and ADNI3 were used, with the following slice counts: PPMI (TrioTim, 41 slices), IXI (Gyroscan Intera: 322, Intera: 185, GE unspecified: 74), SALD (TrioTim: 494), and ADNI3 (Prisma: 69, Prisma Fit: 167). Data were split into 5 folds, with performance averaged across folds.

\subsection{Evaluation Metrics for Generation and Classification}
\label{appendix:eval_metrics}

We evaluate our method against benchmarking methods using metrics designed for both medical image generation and classification. These metrics capture distributional similarity, sample-level fidelity, diversity, and predictive performance.

To rigorously assess the quality of generated medical images, we employ metrics that evaluate both distributional similarity, which measures how closely generated images match real data, and sample-level fidelity and diversity, which measure how realistic and varied the images appear. This dual perspective provides a comprehensive understanding of generative performance in the medical domain.

\subsubsection{Distributional Similarity Metrics.}
\label{subsubsec:distr_sim}
The Fréchet Inception Distance (FID) \citep{heusel2017gans} is a standard metric for comparing real and generated image distributions. It measures the Fréchet distance between multivariate Gaussian distributions fitted to deep features, typically extracted from the pool3 layer of an ImageNet-pretrained InceptionV3 network. However, this approach is suboptimal for medical imaging due to the domain gap between natural and medical images.

To address this, we adopt domain-specific adaptations. FID\textsubscript{Rad} leverages features from an InceptionV3 model pretrained on RadImageNet \citep{mei2022radimagenet}, a large-scale medical imaging dataset. This improves alignment with clinical image characteristics and has been shown to correlate more strongly with radiological quality and anatomical fidelity \citep{fernandez2024generating}. In addition, we employ FID\textsubscript{SwAV} \citep{morozov2021self}, which replaces Inception features with embeddings from SwAV, a self-supervised model trained with clustering and contrastive objectives \citep{caron2020unsupervised}. Unlike ImageNet features, SwAV embeddings are robust to domain shifts and better suited for grayscale medical images, capturing subtle textures and anatomical structures \citep{woodland2024feature}.

Despite its popularity, FID is biased when computed on small sample sizes, a common constraint in medical datasets. This arises because Gaussian parameters are poorly estimated under limited data, resulting in high variance and unstable scores. To mitigate this, we also report the Kernel Inception Distance (KID) \citep{binkowski2018demystifying}, which computes the squared Maximum Mean Discrepancy using polynomial kernels. Unlike FID, KID is an unbiased estimator and therefore more reliable for small datasets. To maintain consistency with domain-specific features, we compute KID\textsubscript{Rad} using RadImageNet-pretrained embeddings.

Finally, to further stabilize estimates, we include bootstrapped versions of the domain-adapted metrics, denoted FID\textsuperscript{b}\textsubscript{Rad} and KID\textsuperscript{b}\textsubscript{Rad}. These are obtained via repeated random subsampling, and we report the 95\% confidence intervals across subsamples. This reduces variance and improves robustness in small-sample settings.

By combining standard FID with domain-adapted, unbiased, and bootstrapped variants, we construct a robust framework for measuring global distributional similarity between real and generated medical images.

\subsubsection{Sample-Level Fidelity and Diversity Metrics.}
\label{subsubsec:sampl_sim}
While distributional metrics capture overall similarity, they do not directly assess sample-level fidelity (how realistic individual images appear) or diversity (how well the generator covers variations in the data). To address this, we employ a complementary set of metrics: Improved Precision and Recall, Density and Coverage (PRDC), and Structural Similarity (MS-SSIM).

Improved \textit{precision (P)} \citep{kynkaanniemi2019improved} measures the fraction of generated samples that lie within the support of real data, estimated via a fixed-radius nearest-neighbor approach in feature space. Higher values (up to 1.0) indicate greater fidelity, meaning generated images are visually indistinguishable from real ones.  
Conversely, \textit{recall (R)} \citep{kynkaanniemi2019improved} measures the fraction of real images that are covered by at least one generated sample, reflecting diversity. Higher values (up to 1.0) suggest the generator successfully captures variations in the real distribution.  
  
To provide a more granular view, we additionally compute \textit{density (D)} \citep{naeem2020reliable}, which counts the average number of generated samples within the neighborhood of each real image using $k$-nearest neighbors. High density (up to 1 in the normalized version) indicates that many generated samples cluster near real images, while low density indicates sparse coverage around real images. Complementing this, \textit{coverage (C)} \citep{naeem2020reliable} quantifies the fraction of real images that are matched by at least one generated image within a fixed distance. Coverage ranges from 0 to 1: higher values indicate that the generator spans almost all distinct regions of the real distribution, while lower values indicate that some modes are missing entirely, even if the generator produces realistic samples elsewhere. Whereas recall emphasizes broad inclusion, coverage specifically evaluates whether the generator touches all distinct regions of the real distribution. Interpreted together, density and coverage provide a complete view of generation quality: high density with low coverage signals realistic samples concentrated in few regions (mode collapse), while high density with high coverage indicates both realistic and diverse generation.

All PRDC metrics are computed in the feature space of the RadImageNet-pretrained InceptionV3 model, ensuring domain-specific alignment with medical image characteristics.  

To directly assess intra-class and inter-class diversity, we use the Structural Similarity Index Measure (SSIM)\citep{wang2004image}, which compares luminance, contrast, and structural information between two images. In particular, we adopt the Multi-Scale SSIM (MS-SSIM)\citep{wang2003multiscale}, which extends SSIM by evaluating image similarity across multiple spatial resolutions, capturing both coarse and fine details. Following prior work~\citep{odena2017conditional, dash2017tac, dragan2023evaluating}, we compute MS-SSIM scores over 500 randomly selected pairs of generated images, distinguishing between intra-class and inter-class diversity due to our conditional setting. Intra-class diversity ($\text{MS-SSIM}^\text{intra}$) is computed among samples of the same class. Lower values indicate greater diversity within a class, while higher values suggest mode collapse toward visually similar outputs. Inter-class diversity ($\text{MS-SSIM}^\text{inter}$) is computed across samples from different classes. Lower values reflect good class separability, whereas higher values may indicate insufficient differentiation between categories. 

\subsubsection{Classification Metrics.}
\label{app:class_metrics}
For classification models, we report standard metrics including overall Accuracy, Balanced Accuracy, Area Under the ROC Curve (AUC), and macro-averaged Precision, Recall, and F1-score. Accuracy measures the fraction of correctly classified samples, while Balanced Accuracy accounts for class imbalance by averaging recall across all classes. AUC evaluates the model's discriminative ability across decision thresholds. Macro-averaged Precision, Recall, and F1-score summarize per-class performance and provide a comprehensive view of sensitivity and specificity across categories.

\subsection{Benchmark Models for Comparison}
\label{appendix:benchmarks}

To evaluate the effectiveness of MAGIC-Flow, we benchmark it against a diverse set of competitive state-of-the-art models spanning adversarial training, diffusion-based generation, and latent-variable modeling. These baselines are well-established in the literature and cover a broad methodological spectrum, ensuring a robust and informative comparison.

\subsubsection{Generative Models}
\label{appendix:benchmarks_gen}
\begin{itemize}
    \item \textbf{SNGAN}: Spectral Normalization GAN \citep{miyato2018spectral} introduces spectral normalization to the discriminator’s weights, enforcing a Lipschitz constraint that stabilizes training. It is a widely adopted conditional GAN known for generating high-fidelity samples with reliable convergence.
    \item \textbf{StyleGAN2-DiffAug-LeCam}: This approach extends StyleGAN2 \citep{karras2020analyzing} by incorporating DiffAugment \citep{zhao2020differentiable} and LeCam regularization \citep{tseng2021regularizing}, which together improve training efficiency and generalization on limited data. These augmentations help maintain strong conditional generation performance without requiring large-scale datasets.
    \item \textbf{ADC-GAN}: The Auxiliary Discriminative Classifier GAN \citep{hou2022conditional} enhances the BigGAN framework by integrating an auxiliary classifier that simultaneously predicts class labels and discriminates real versus fake samples via class-specific labeling. This dual role improves intra-class diversity and promotes stable training dynamics compared to traditional conditional GAN variants like AC-GAN \citep{odena2017conditional}.
    \item \textbf{DDPM}: Denoising Diffusion Probabilistic Models \citep{nichol2021improved, dhariwal2021diffusion} are non-adversarial generative models that learn to reverse a fixed noising process through iterative denoising. We adopt the conditional DDPM framework with classifier guidance, which leverages a pretrained classifier’s gradients during sampling to steer generation towards the desired class, significantly improving conditional sample quality and fidelity.
    \item \textbf{CVAE}: The Conditional Variational Autoencoder \citep{sohn2015learning} models conditional distributions via latent variables by injecting conditioning information into both the encoder and decoder networks. This probabilistic approach captures multimodal outputs and enables principled maximum likelihood training.
\end{itemize}

\subsubsection{Discriminative Models for Classification}
\label{appendix:benchmarks_class}
To establish a robust comparison framework, we further benchmark against convolutional neural networks (CNNs) pretrained on RadImageNet, as well as Vision Transformers (ViTs).  

\textbf{CNNs.} RadImageNet is a large-scale medical imaging database containing approximately 1.35 million annotated CT, MRI, and ultrasound images across 11 anatomic regions and 165 pathologic labels \citep{mei2022radimagenet}. Unlike ImageNet pretraining on natural images, RadImageNet provides domain-specific initialization that has been shown to improve transferability to radiologic tasks. We selected four widely used CNN architectures, all initialized with RadImageNet-pretrained weights and subsequently fine-tuned on our dataset:  
\begin{itemize}
    \item \textbf{ResNet-50} \citep{he2016deep} -- A residual network that mitigates vanishing gradients with skip connections, providing a strong balance of depth and efficiency.
    \item \textbf{DenseNet-121} \citep{huang2017densely} -- A densely connected network that enhances feature reuse and parameter efficiency.
    \item \textbf{InceptionV3} \citep{szegedy2016rethinking} -- An architecture employing factorized convolutions and dimensionality reduction to capture multi-scale features efficiently.
    \item \textbf{InceptionResNetV2} \citep{szegedy2017inception} -- A hybrid combining Inception modules with residual connections for deeper feature extraction and stable optimization.
\end{itemize}

\textbf{Vision Transformers.} Unlike CNNs, which learn spatial hierarchies through convolutions, ViTs partition images into fixed-size patches that are linearly projected and processed by transformer encoder layers with self-attention \citep{dosovitskiy2020image}. Prior work has demonstrated that ViTs pretrained on large datasets (e.g., ImageNet-21k) can achieve strong performance in medical imaging tasks, particularly when combined with CNN backbones \citep{dosovitskiy2020image, jain2024comparative}. We include the following variants:
\begin{itemize}
    \item \textbf{ViT-v1/32} -- A baseline transformer dividing $224 \times 224$ images into $32 \times 32$ patches.
    \item \textbf{ViT-ResNet/16} -- A hybrid model that extracts features with a ResNet backbone before transformer encoding. It uses $16 \times 16$ patches and ImageNet-21k pretraining, enabling improved initialization and faster convergence relative to standalone ViTs.
    \item \textbf{Swin Transformer} \citep{liu2021swin} -- A hierarchical vision transformer that introduces shifted window attention to efficiently model both local and global dependencies. Swin achieves state-of-the-art performance on natural image benchmarks such as ImageNet, COCO, and ADE20K, and has been shown to transfer well to medical imaging tasks by capturing fine-grained features while maintaining scalability \citep{liu2021swin, he2023transformers}.
\end{itemize}

Together, these CNN and ViT baselines provide strong and diverse comparators for evaluating the performance of MAGIC-Flow in medical image classification.

\end{document}